\pdfoutput=1

\documentclass[11pt]{article}

\usepackage[final]{acl}
\usepackage{booktabs}
\usepackage{float}
\usepackage{amsmath}
\usepackage{multirow}
\usepackage{lipsum}
\usepackage{amssymb}
\usepackage{tabularx} 
\usepackage{listings}
\usepackage{xcolor}
\usepackage{caption}
\usepackage{tcolorbox}

\lstdefinestyle{pythonstyle}{
    language=Python,
    backgroundcolor=\color{gray!10},
    basicstyle=\ttfamily\small,
    keywordstyle=\color{blue},
    stringstyle=\color{green!50!black},
    commentstyle=\color{red!50!black},
    breaklines=true,
    showstringspaces=false
}

\usepackage{times}
\usepackage{latexsym}

\usepackage[T1]{fontenc}

\usepackage[utf8]{inputenc}

\usepackage{microtype}

\usepackage{inconsolata}

\usepackage{booktabs}
\usepackage{multirow}
\usepackage{graphicx}

\usepackage{xspace}

\usepackage{xcolor}
\usepackage{siunitx}
\usepackage{makecell}

\definecolor{mydarkgreen}{RGB}{0,100,0}
\newcommand{\up}[1]{\textcolor{mydarkgreen}{\(\uparrow\,#1\)}}
\newcommand{\down}[1]{\textcolor{red}{\(\downarrow\,#1\)}}

\newcommand{\method}{\textsc{CIE}\xspace}
\newcommand{\metric}{\textsc{CPR}\xspace}
\newcommand{\dataset}{\textsc{VerbosityCTRL}\xspace}

\title{\method: Controlling Language Model Text Generations Using Continuous Signals}

\author{
 \textbf{Vinay Samuel\textsuperscript{1,2,}\footnotemark}\\ \quad
 \textbf{Harshita Diddee\textsuperscript{2}} \quad
 \textbf{Yiming Zhang\textsuperscript{2}} \quad
 \textbf{Daphne Ippolito\textsuperscript{2}} \quad
\\ \\
 \textsuperscript{1}University of Maryland, College Park,
 \textsuperscript{2}Carnegie Mellon University 
\\
}

\begin{document}
\maketitle

\begin{abstract}

Aligning language models (LMs) with user intent is becoming increasingly relevant to enhance user experience.
This calls for designing methods that can allow users to control the properties of the language that LMs generate, for example, controlling the length of the generation or the complexity of the language that gets chosen.
Most existing work attempts to integrate users' control by conditioning LM generations on natural language prompts or discrete control signals, which are often brittle and hard to scale.
In this work, we are interested in \textit{continuous} control signals, ones that exist along a spectrum that can't easily be captured in a natural language prompt or via existing techniques in conditional generation.
Through a case study in controlling the precise response-length of generations, we demonstrate how an LM can be finetuned to expect a control vector that is interpolated between a ``low'' and a ``high'' token embedding.
Our method more reliably exerts response-length control than in-context learning methods or fine-tuning methods that represent the control signal as a discrete signal.

\end{abstract}
\renewcommand{\thefootnote}{\fnsymbol{footnote}}
\footnotetext[0]{* Correspond to vsamuel@umd.edu.}
\renewcommand{\thefootnote}{\arabic{footnote}}

\section{Introduction}

\begin{figure}[!t]
  \centering
  \includegraphics[width=\columnwidth]{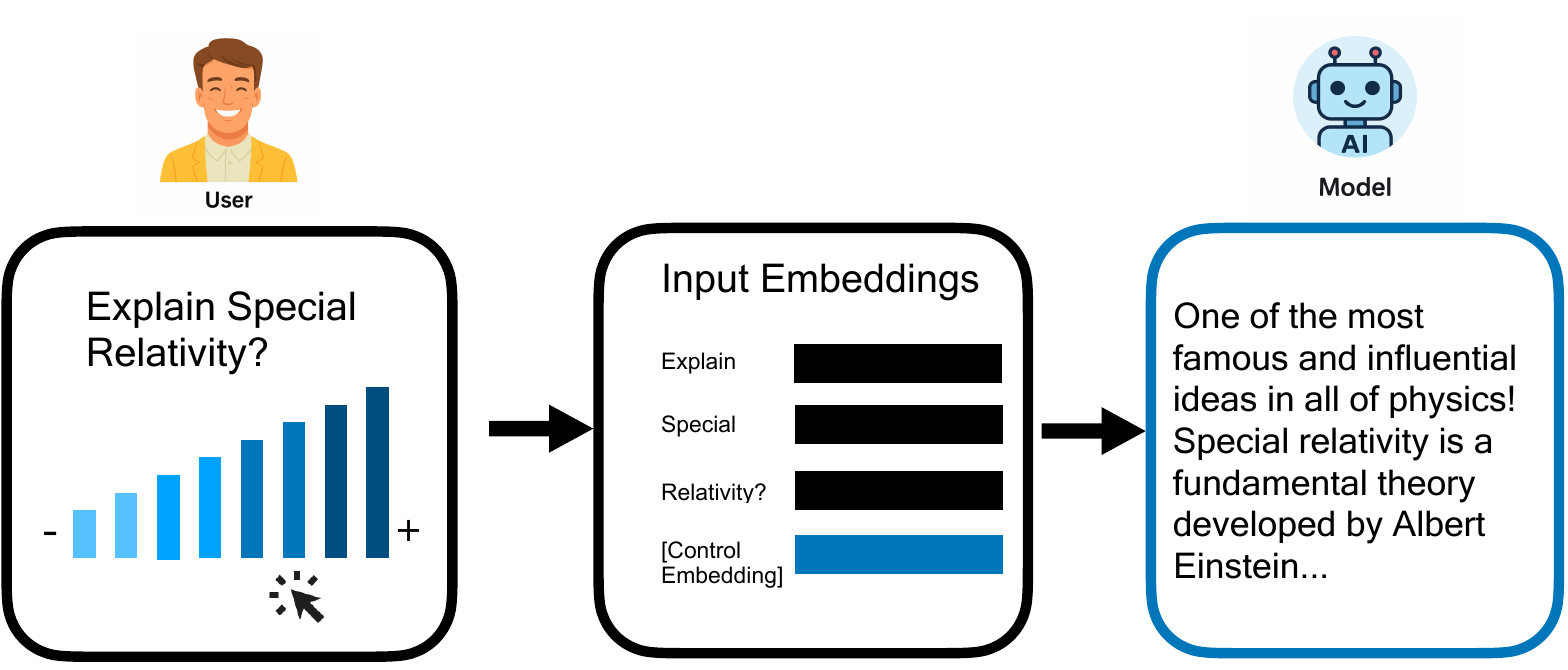}
  \caption{When using \method for controlling response length, the user inputs both an instruction as well as a desired answer length.
  A control embedding for this response length is interpolated between the lower and upper bound control embeddings which were learned during training.
  The control embedding is appended to the input token embedding sequence of the instruction, and the LM which has been finetuned to expect this input generates an attribute-controlled response.}
  \label{fig:teaser}
\end{figure}

Instruction-tuned language models have demonstrated remarkable capabilities in generating coherent responses to user instructions \citep{alpaca, InstructGPT, rewritelm}. 
However, users often want to influence specific properties of the generated text beyond just the content--for example, controlling the length of the generation (as illustrated in Figure~\ref{fig:teaser}), the complexity of the language, the sentiment, or the tone.

There are many approaches to controlling such attributes in language model outputs.
While discrete control signals (such as special tokens or words that get added to the user's prompt) have shown promise \citep{style, rlprompt, ctrl, plugandplay}, they have inherent limitations.
Discrete approaches struggle to provide fine-grained control (especially for properties that exist on a continuous spectrum), and they can require extensive training to achieve competence \citep{ctrl, ruler}. 
Meanwhile, methods that achieve controllability through natural-language instructions tend to be brittle, with small, immaterial changes to the verbalization potentially leading to inconsistent results \citep{prosa}.

In this work, we focus on \textit{continuous} control signals that exist along a spectrum rather than discrete tokens or natural language prompts.
We introduce \textbf{C}ontrol through \textbf{I}nterpolated \textbf{E}mbeddings (\method), a method enabling nuanced control over text generation through conditioning with a single control embedding.

While similar approaches have demonstrated effectiveness in domains such as strength-conditioning in chess \citep{allie}, we extend this methodology to precisely control specified attributes in language model outputs.
Using continuous embeddings for incorporating continuous control signals rather than natural language instructions 
allows us to circumvent LLMs' limited numerical understanding \citep{number}.
Our approach also offers advantages over discrete control token methods by allowing smoother transitions between attribute intensities and more precise steering of generative outputs.

The core mechanism of \method is a \textit{control embedding} that is appended to the token embeddings of the user instruction.
We augment the LM's existing embedding matrix with two new embeddings corresponding to the lowest and highest possible values for a given attribute.
During fine-tuning and inference, the control embedding is computed through linear interpolation between these low and high embeddings.
Given a dataset of instruction-answer pairs annotated with control values, \method enables LMs to learn a mapping from the control embedding's position to the attribute's degree of control.

To demonstrate our proposed method \method's effectiveness, we apply it to response length control (as measured by word count) and compare our results with both a prompting baseline and a state-of-the-art discrete signal approach \citep{ruler}.
\footnote{Our open-source code and training data are available at \href{https://github.com/vsamuel2003/CIE}{https://github.com/vsamuel2003/CIE}.}

\section{Method}
To produce coherent text while adhering to specified attributes, \method creates a positional mapping between control embedding locations in the input embedding and the specific attribute $a$ being controlled. This mapping enables the language model to generate text conditioned on desired attributes. On a parameter level, the only additional parameters that \method introduces to the LM are a control embedding matrix $\mathbf{E} \in \mathbb{R}^{2 \times D}$ where D is the embedding dimension of the model.

The control embedding matrix is meant to represent the embedding vectors for lower and upper bound values of the given attribute. For a given conditioned value $c$,
we define the control embedding matrix $\mathbf{E}$ as $$\mathbf{E} =
\begin{pmatrix}
\mathbf{e}_{\text{lower}} \\
\mathbf{e}_{\text{upper}}
\end{pmatrix} \in \mathbb{R}^{2 \times D}$$ where $\mathbf{e}_{\text{lower}}$ represents the embedding vector for the lower bound of control values for $a$ in the training data $c_{\text{lower}}$ and $\mathbf{e}_{\text{upper}}$ is the embedding vector for the upper bound of the allowed values of $a$ in the training data $c_{\text{upper}}$. For a given $c$, we calculate the control embedding vector for $c$ as $\mathbf{e}_{\text{c}} = \alpha \mathbf{e}_{\text{lower}} + (1 - \alpha)\mathbf{e}_{\text{upper}}$ where $\alpha = \frac{c_{\text{upper}} - c}{c_{\text{upper}} - c_{\text{lower}}}$. 

A \texttt{<control-embedding>} token is appended to instructions without expanding vocabulary. During the forward pass, control positions are replaced with interpolated embeddings computed from two learned embeddings ($\mathbf{e}_{\text{lower}}$ and $\mathbf{e}_{\text{upper}}$) before transformer processing. The interpolated control embedding provides a continuous conditioning signal that influences all subsequent transformer layers. Unlike discrete tokens that compete for attention with content tokens, the control embedding acts as a persistent bias that guides generation decisions throughout the decoding process.

The control embedding is interpolated between trained bounds and injected into instruction embeddings for standard autoregressive generation. During decoding, the control embedding establishes a representation space bias that influences next-token probability distributions. The learned embeddings encode attribute-specific patterns that, when interpolated, provide fine-grained control over generation characteristics.

The training dataset $\mathcal{D}$ comprises triplets of $(i,a,wc)$, where $i$ represents the instruction, $a$ denotes the ground truth answer, and $wc$ indicates the conditioned response length. During training, $wc$ is set to match the response length in $a$. We establish engineering-determined lower and upper bounds ($c_{\text{lower}}$ and $c_{\text{upper}}$) for the controlled attribute, with all $wc$ values clamped within this range before calculating $e_c$. Following the approach of \citet{allie}, we curate $\mathcal{D}$ to maintain an approximately uniform distribution of response lengths, ensuring sufficient training for $\mathbf{e}_{\text{lower}}$ and $\mathbf{e}_{\text{upper}}$.

The method adds only a $(2 \times D)$ control embedding matrix to the base model, requiring minimal modifications to standard causal language modeling.

\section{Experiments}
To demonstrate the efficacy of continuous control signals, we apply \method to the attribute of response-length. This section details our experimental setup and results. We include details regarding our training and validation setups, including hardware and hyperparameters, in Appendix~\ref{sec:training} and Appendix~\ref{sec:validation}, respectively.

\subsection{Datasets}
To evaluate how effectively \method controls response-length in text generation, we introduce \dataset, which combines data from both conversational and traditional NLP style datasets namely: MSMarco \citep{msmarco}, OpenAssistant 1/2 \citep{oa}, and Databricks Dolly 15k \citep{dolly}. We divide this dataset into \dataset$_{\text{train}}$ and \dataset$_{\text{val}}$. Additionally, we created \dataset$_{\text{range}}$, an augmented version of \dataset$_{\text{val}}$ where each instruction appears in 10 variations with target \texttt{word\_count} values from 20 to 200 in 20-unit increments. For out-of-distribution validation, we utilized the Alpaca-LI dataset \citep{lift}. Complete data processing methodology and dataset statistics are available in Appendix~\ref{sec:dataset_processing} and Appendix~\ref{sec:data_stats}, respectively.

\subsection{Metrics}
We evaluate our results using the Conditioning Precision Ratio (\metric) metric, which measures how accurately models follow response length instructions. \metric simply checks for an exact match between the generated response length and the conditioned response length, returning 1 for a perfect match and 0 otherwise. We also define \metric@k, which introduces a relative tolerance factor, accepting response lengths within $\pm k$ percent of the target (e.g., \metric@0.1 allows deviations within 10\% of the specified response length). To verify that \method preserves language generation quality, we calculate win-rates between model outputs and a prompt baseline, counting both wins and ties as evidence that \method does not harm generation abilities. Additional details on this evaluation approach are available in Appendix~\ref{sec:win_rates}.

\subsection{Models}
We finetune and evaluate several open-source LLMs namely LLaMA-3-8b-Instruct \citep{llama}, gemma-7B-it \citep{gemma}, and Qwen1.5-7B-Chat \citep{qwen1.5}. These particular models were selected to show the efficacy of \method while also enabling fair comparison to the results from \citet{ruler}.  

\subsection{Baselines}
We compare \method to a prompting baseline where we prepend the sentence \texttt{``Respond to the following in exactly \{wordcount\} words.''} where \texttt{wordcount} is replaced with the conditioned response length. This prompt was chosen as the highest performing prompt across three prompts that were tested. Details regarding all prompt experiments, as well as their performances, are included in Appendix~\ref{sec:baseline_prompt}. 

In addition, we also compare to the Ruler \citep{ruler} method, which uses Meta Length Tokens, a discrete token embedding for unique response-length directive provided by the user.

\section{Results}

\begin{table*}[!h]
  \centering
  \resizebox{\textwidth}{!}{%
    \begin{tabular}{l|ccc|ccc|ccc}
      \toprule
      \multirow{2}{*}{\textbf{Model}}
        & \multicolumn{3}{c|}{\textbf{\dataset$_\text{val}$}}
        & \multicolumn{3}{c|}{\textbf{\dataset$_\text{range}$}}
        & \multicolumn{3}{c}{\textbf{Alpaca-LI}} \\
      \cmidrule(lr){2-10}
        & \metric        & \metric @0.05     & \metric @0.1
        & \metric        & \metric @0.05     & \metric @0.1
        & \metric        & \metric @0.05     & \metric @0.1 \\
      \midrule

      \makecell[l]{LLaMA-3-8B-IT\\LLaMA-3-8B-IT$_{\method}$}
        & \makecell[l]{9.80\\9.50\down{0.30}}
        & \makecell[l]{22.70\\45.80\up{23.10}}
        & \makecell[l]{38.90\\72.70\up{33.80}}
        & \makecell[l]{3.19\\5.05\up{1.86}}
        & \makecell[l]{17.40\\39.08\up{21.68}}
        & \makecell[l]{39.04\\71.65\up{32.61}}
        & \makecell[l]{5.43\\9.04\up{3.61}}
        & \makecell[l]{21.95\\45.02\up{23.07}}
        & \makecell[l]{43.67\\72.40\up{28.73}} \\

      \midrule

      \makecell[l]{gemma-7B-IT\\gemma-7B-IT$_{\method}$}
        & \makecell[l]{0.80\\8.90\up{8.10}}
        & \makecell[l]{4.90\\28.90\up{24.00}}
        & \makecell[l]{8.80\\47.50\up{38.70}}
        & \makecell[l]{0.73\\4.42\up{3.69}}
        & \makecell[l]{3.87\\37.90\up{34.03}}
        & \makecell[l]{9.94\\68.37\up{58.43}}
        & \makecell[l]{1.58\\4.75\up{3.17}}
        & \makecell[l]{9.05\\40.95\up{31.90}}
        & \makecell[l]{17.19\\65.16\up{47.97}} \\

      \midrule

      \makecell[l]{Qwen-1.5-7B\\Qwen-1.5-7B$_{\method}$}
        & \makecell[l]{4.60\\9.80\up{5.20}}
        & \makecell[l]{9.40\\39.80\up{30.40}}
        & \makecell[l]{16.60\\67.80\up{51.20}}
        & \makecell[l]{0.91\\4.77\up{3.86}}
        & \makecell[l]{5.96\\31.55\up{25.59}}
        & \makecell[l]{14.82\\62.71\up{47.89}}
        & \makecell[l]{2.94\\4.20\up{1.26}}
        & \makecell[l]{8.82\\20.36\up{11.54}}
        & \makecell[l]{17.42\\38.46\up{21.04}} \\

      \bottomrule
    \end{tabular}%
  }
  \caption{Results of prompt baseline and \method on \dataset$_\text{val}$, \dataset$_\text{range}$, Alpaca-LI. We present results on the \metric, \metric@0.05 and \metric@0.1 metric to present results of various competitive thresholds.}
  \label{tab:main_results}
\end{table*}

\begin{figure}[t]
  \centering
  \includegraphics[width=\linewidth]{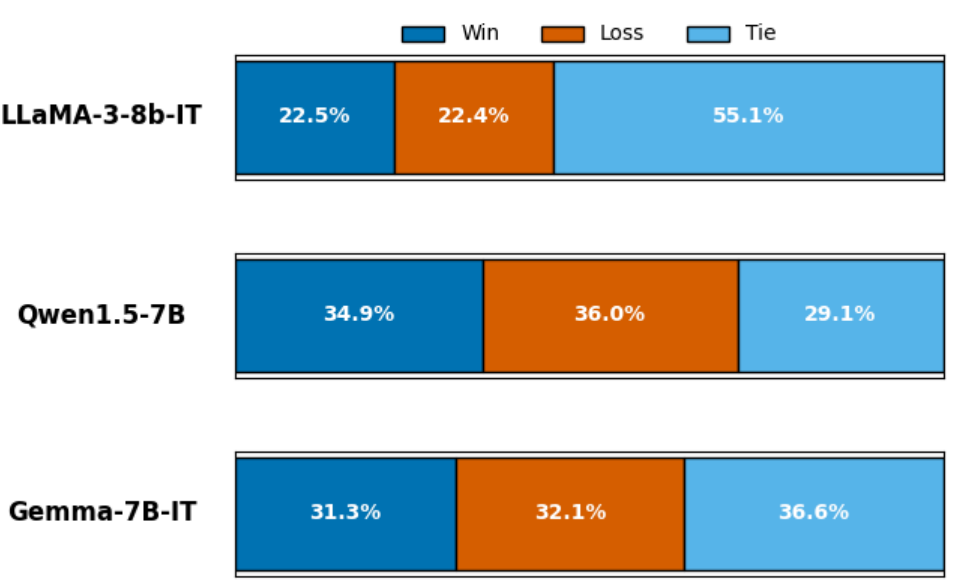}
  \caption{Win rates of \method models vs a length conditioned prompting baseline as defined in Appendix~\ref{sec:win_rates}. The judge model used was GPT-4.}
  \label{fig:win_rates}
\end{figure}

Table~\ref{tab:main_results} presents our main findings of comparing models trained using \method to the prompt baseline.

\paragraph{Enhanced Performance with \method.}
Our \method approach significantly improved response length control across all evaluated models and datasets. On \dataset$_{\text{val}}$, \method increased average \metric from 0.80-9.80\% to 8.90-9.80\%, with substantial gains for initially weaker models like gemma-7B-IT (0.80\% to 8.90\%). The most dramatic improvements occurred at the \metric@0.05 threshold, with increases of 23.10, 24.00, and 30.40 percentage points for LLaMA-3-8B-IT, gemma-7B-IT, and Qwen-1.5-7B respectively. Similar patterns emerged on \dataset$_{\text{range}}$, with average \metric rising from 1.6\% to 4.8\%. These improvements generalized to out-of-distribution data (Alpaca-LI), confirming \method's cross-contextual robustness. We observed an inverse relationship between baseline capability and improvement magnitude—weaker models achieved more dramatic relative gains, with gemma-7B-IT showing over 10$\times$ improvement at exact-match \metric. Stronger models like LLaMA-3-8B-IT made minor sacrifices at exact-match for substantial gains at relaxed thresholds. The diminishing relative gains at \metric@0.1 suggest base models already capture many approximate matches at looser tolerances. Detailed analyses are available in Appendices~\ref{sec:detailed_analysis} and~\ref{sec:diff_ranges}.

\paragraph{Win-rates to gauge coherent language generation capabilities.} 
Figure~\ref{fig:win_rates} shows win-rates between \method and a response length conditioned prompt baseline for each model. We consider both wins and ties as signals that language generation and instruction-following capabilities remain intact. Across all models, combined win and tie rates exceed loss rates, indicating \method generally preserves these capabilities. A more detailed analysis is present in Appendix~\ref{sec:win_rate_analysis}.

\paragraph{Comparison to RULER.}
Table~\ref{tab:ruler} presents comparisons between \method and \textsc{Ruler}, where both were trained using \dataset$_{\text{train}}$ and evaluated on a validation set in-distribution to that used in \citet{ruler}. We compare only on \textsc{Ruler}'s level 0 (1-150 words), as it aligns with our experimental response length ranges. We adopt the same metrics: \texttt{Precise Match (PM)}, allowing $\pm$10 words at all lengths, and \texttt{Flexible Match (FM)}, allowing $\pm$10 words for lengths below 80 and $\pm$20 words for lengths greater than or equal to 80.

\method consistently outperforms \textsc{Ruler} across all model variants. For gemma-7B-IT, \method achieves absolute improvements of 9.09 and 7.86 points in PM and FM scores, respectively. LLaMA-3-8B shows even more substantial gains of 16.30 and 16.18 points, while Qwen1.5-7B demonstrates improvements of 6.54 and 6.44 points. 

To understand why CIE outperforms discrete methods, we conducted scaling experiments training LLaMA-3-8B on both RULER and CIE at 25\%, 50\%, 75\%, and 100\% of training data. Results from Table~\ref{tab:scaling_analysis} show CIE maintains superiority across all data scales without convergence. CIE's continuous interpolation creates a structured embedding space where similar lengths have similar representations, enabling generalization between target values with fewer examples per length. The continuous space inherently captures ordinal relationships between control values, whereas discrete methods lack this geometric structure. Additionally, the requirement of adding in additional MLT tokens in the RULER method makes the method non-trivial to scale to larger word counts.

These significant performance gains highlight \method's effectiveness in enhancing precision response length conditioning, suggesting it provides a more robust framework for output length control compared to state-of-the-art methods.

\begin{table}[h]
\centering
\tiny
\begin{tabular}{l|c|c|c|c}
\hline
\textbf{Model} & \textbf{25\%} & \textbf{50\%} & \textbf{75\%} & \textbf{100\%} \\
\hline
LLaMA-3-8B\_Ruler & 11.80/14.20 & 24.20/28.00 & 38.00/42.50 & 49.22/53.33 \\
LLaMA-3-8B\_CIE & 13.08/16.32 & 26.72/30.38 & 41.46/44.79 & 65.52/69.51 \\
\hline
\end{tabular}
\caption{Performance Analysis: Why CIE Outperforms Discrete Methods. The results show Precise Math / Flexible Match}
\label{tab:scaling_analysis}
\end{table}

\begin{table}
\centering
\small
\begin{tabular}{ccc}
\toprule
Model & PM & FM \\
\midrule
gemma-7B-IT & 15.52 & 18.85 \\
gemma-7B$_{\text{Ruler}}$ & 62.42 & 67.52 \\
gemma-7B$_{\text{\method}}$ & \textbf{71.51} & \textbf{75.38} \\
\midrule
LLaMA-3-8B-Instruct & 34.59 & 40.02 \\
LLaMA-3-8B$_{\text{Ruler}}$ & 49.22 & 53.33 \\
LLaMA-3-8B$_{\text{\method}}$ & \textbf{65.52} & \textbf{69.51} \\
\midrule
Qwen1.5-7B-Chat & 24.28 & 27.38 \\
Qwen1.5-7B$_{\text{Ruler}}$ & 39.91 & 44.79 \\
Qwen1.5-7B$_{\text{\method}}$ & \textbf{46.45} & \textbf{51.23} \\
\bottomrule
\end{tabular}
\caption{Performance comparison between prompt-based length condition, \textsc{Ruler}, and \method on a validation set in distribution with the validation set used in \textsc{Ruler}. Both the \textsc{Ruler}, and \method models were trained on \dataset$_{\text{train}}$}
\label{tab:ruler}
\end{table}


\section{Related Works}
\paragraph{Related Work}
Prior work on attribute control in LLMs can be categorized into three approaches. \textit{Discrete control signals} include \citet{ctrl}'s conditional transformer with control codes for style and content steering, \citet{plugandplay}'s combination of pre-trained LMs with attribute classifiers that guide generation without fine-tuning, and \citet{ruler}'s "Meta Length Tokens" for controlling word counts within specified deviation levels. \textit{Prompt engineering} approaches comprise \citet{ramp}'s retrieved examples with special markings for translation attribute control, \citet{magic}'s "magic words" that steer models toward specific outputs, and \citet{lift}'s fine-tuning with templates that condition for desired word counts. \textit{Continuous control signals} include \citet{tailor}'s soft-prompt tuning where trainable embedding vectors guide frozen LMs toward target attributes, \citet{mixture}'s learned prompt mixtures for multiple attribute constraints, and \citet{concept}'s approach of steering generation via concept vectors in the LLM's hidden activation space.

\section{Conclusion}
Fine-grained control over language model outputs represents a critical capability for deploying these systems in contexts requiring adaptable text generation. While the vast majority of current controllability approaches are based on discrete signals, we believe that the fine-grained controllability of continuous signals is necessary for the evolving user demands of conversational LMs. Through our proposed method, we have established a framework that effectively modulates response length while preserving content fidelity. 

Beyond word count, future work should aim to apply CIE as well as other continuous signal approaches to control concrete attributes like sentence count, character count, and complexity measures. While subjective properties like toxicity and bias present additional challenges in defining meaningful continuous ranges, as discussed in the Limitations of this paper, they represent important directions for future exploration.

\section*{Limitations}
While CIE provides a general framework extendable to any attribute it may be sample efficient for generation attributes that lack explicit control signals over a wide continuous range; For instance, controlling the `politeness' \cite{Yin2024ShouldWR} or `sarcasm' \cite{Zhang2024SarcasmBenchTE} requires creating a meaningful continuous range for notions of a more `polite' response versus a less `polite' response which may need thoughtful reconciliation of fairly subjective notion of what signals identify examples in either category.

\section*{Acknowledgments}
The authors acknowledge AI assistance as follows: Several sections underwent refinement using AI, where the AI was instructed to 'rewrite in clear, coherent, and concise academic style writing while not altering the major points in the provided writing.' All AI-generated content was thoroughly reviewed by the authors before inclusion in the manuscript, ensuring full compliance with ACL ARR guidelines. Additionally some figure components (such as the model and user avatars) were either AI generated or AI refined. We also thank Barry Wang (CMU CSD PhD) for major inputs in designing the teaser figure.

\bibliography{custom}

\appendix

\section{Dataset Processing}
\label{sec:dataset_processing}
Data filtration was conducted to remove all non-english instances as well as coding instances. Furthermore \dataset includes a \texttt{word\_count} field that is determined based on the word count of the ground truth \texttt{answer} for each data instance. We conduct several data processing steps to our data in \dataset to ensure high quality. Firstly, to maintain a uniform distribution of word counts in bins of size 25, we limit our data to those with word counts between 1 and 200 words, where Listing~\ref{lst:word_count} shows our function for counting words, which is the same as \citet{ruler}. 

\begin{lstlisting}[style=pythonstyle, caption={Word counting function using NLTK}, label={lst:word_count}]
from nltk.tokenize import word_tokenize
import string
def count_words(text):
    return len([word for word in word_tokenize(text) if word not in string.punctuation])
\end{lstlisting}

We additionally conduct a filtering to remove all instances of non-English instructions/answers as well as datapoints that contain coding keywords. Our approach to filtering these datapoints is shown in Listing~\ref{lst:filtering}

\begin{lstlisting}[style=pythonstyle, caption={Non-English and coding instances filteringh}, label={lst:filtering}]
import langdetect

# Function to check for programming terms
def has_programming_terms(text):
    keywords = ["java", "python", "c++", "def", "return", "program", "function", "script", "html", "css", 
                "javascript", "php", "sql", "ruby", "swift", "kotlin", "go", "rust", "scala", "haskell", 
                "erlang", "elixir", "dart", "typescript", "c#", "visual basic", "objective-c", "assembly", 
                "matlab", "perl", "shell", ".js", "json", "xml", "<", ">", "lorem ipsum", "\document", "---", "excel", 
                "https", "tabular", "\end", "ascii", "*", "translate", "korean", "IP"]
    text_lower = text.lower()
    return any(keyword in text_lower for keyword in keywords)

# Function to check if text is in English
def is_english(text):
    try:
        return langdetect.detect(text) == 'en'
    except:
        return False
\end{lstlisting}

\section{Baseline Prompts}
\label{sec:baseline_prompt}

\begin{table}[!h]
\small                                         
\renewcommand{\arraystretch}{1.1}              
\setlength{\tabcolsep}{4pt}                    
\centering
\begin{tabularx}{\linewidth}{|>{\centering\arraybackslash}p{2cm}|X|}
  \hline
  \textbf{Label} & \textbf{Prompt} \\ \hline
  Prompt 1 & The response should have a word count of \texttt{\{wordcount\}}. \\ \hline
  Prompt 2 & The answer should be \texttt{\{wordcount\}} words. \\ \hline
  Prompt 3 & Respond to the following in exactly \texttt{\{wordcount\}} words. \\ \hline
\end{tabularx}
\caption{Word-count control prompts used in our experiments.}
\label{tab:wordcount_prompts}
\end{table}

Due to the brittleness of prompting-based approaches to controlling LM outputs, we experiment with three different prompt templates (see Table~\ref{tab:wordcount_prompts}) that are appended to the beginning of the instruction for data instances as part of our prompt baseline. Prompt 1 is taken from \citet{lift}, prompt 2 is taken from \citet{ruler}, and the authors of this paper created prompt 3. We include a ``best'' performing prompt baseline where for a given instruction $j$ this baseline selects the response from prompt $j \in [1,2,3]$ such that the prediction from prompt $j$ has the closest \metric to 1 for instruction $j$.

\begin{table*}[!h]
  \scriptsize
  \centering
  \begin{tabular}{llccc|ccc|ccc}
    \toprule
    \multirow{2}{*}{\textbf{Model}}
      & \multirow{2}{*}{\textbf{Baseline}}
      & \multicolumn{3}{c|}{\textbf{\dataset$_\text{val}$}}
      & \multicolumn{3}{c|}{\textbf{\dataset$_\text{range}$}}
      & \multicolumn{3}{c}{\textbf{Alpaca-LI}} \\
    \cmidrule(lr){3-5} \cmidrule(lr){6-8} \cmidrule(lr){9-11}
      &
      & \metric   & \metric@0.05 & \metric@0.1
      & \metric   & \metric@0.05 & \metric@0.1
      & \metric   & \metric@0.05 & \metric@0.1 \\
    \midrule
    \multirow{5}{*}{\textbf{LLaMA-3-8B-IT}}
      & Prompt 1   & 3.60      & 9.90       & 18.10
                   & 1.00      & 9.52       & 21.18
                   & 4.07      & 24.89      & 45.02 \\
      & Prompt 2   & 8.00      & 21.80      & 36.60
                   & 2.86      & \underline{18.60} & 37.71
                   & \underline{6.11} & \underline{26.24} & \underline{49.77} \\
      & Prompt 3   & \underline{9.80} & \underline{22.70} & \underline{38.90}
                   & \underline{3.19} & 17.40      & \underline{39.04}
                   & 5.43      & 21.95      & 43.67 \\
      & Random     & 7.30      & 17.70      & 31.50
                   & 2.35      & 15.61      & 32.50
                   & 5.66      & 25.79      & 48.64 \\
      & Best       & \textbf{12.80} & \textbf{37.10} & \textbf{57.90}
                   & \textbf{6.44} & \textbf{34.09} & \textbf{58.31}
                   & \textbf{10.86} & \textbf{48.19} & \textbf{69.46} \\
    \midrule
    \multirow{5}{*}{\textbf{gemma-7B-IT}}
      & Prompt 1   & 0.70      & 4.90       & \underline{9.80}
                   & 0.73      & \underline{5.38} & \underline{12.23}
                   & 0.45      & \underline{11.09} & \underline{23.08} \\
      & Prompt 2   & \underline{0.80} & 5.00       & 9.20
                   & \underline{0.88} & 4.58       & 11.53
                   & 0.68      & 7.47       & 16.52 \\
      & Prompt 3   & \underline{0.80} & 4.80       & 8.80
                   & 0.73      & 3.87       & 9.94
                   & 1.58      & 9.05       & 17.19 \\
      & Random     & 0.70      & \underline{5.20} & 9.60
                   & 0.68      & 4.39       & 11.37
                   & \underline{1.81} & 10.18      & 19.23 \\
      & Best       & \textbf{2.20} & \textbf{11.90} & \textbf{20.00}
                   & \textbf{2.19} & \textbf{11.71} & \textbf{25.06}
                   & \textbf{2.49} & \textbf{20.14} & \textbf{35.07} \\
    \midrule
    \multirow{5}{*}{\textbf{Qwen-1.5-7B}}
      & Prompt 1   & 0.20      & 3.20       & 9.20
                   & 0.39      & 3.71       & 9.04
                   & 2.04      & 9.50       & 16.97 \\
      & Prompt 2   & 2.70      & 5.90       & 11.20
                   & 0.69      & 5.15       & 12.52
                   & 2.49      & \underline{10.41} & \underline{21.27} \\
      & Prompt 3   & \underline{4.60} & \underline{9.40} & \underline{16.60}
                   & \underline{0.91} & \underline{5.96} & \underline{14.82}
                   & \underline{2.94} & 8.82       & 17.42 \\
      & Random     & 2.60      & 7.00       & 12.90
                   & 0.54      & 5.04       & 11.91
                   & 2.71      & 10.18      & 19.46 \\
      & Best       & \textbf{5.50} & \textbf{15.60} & \textbf{29.10}
                   & \textbf{1.91} & \textbf{12.87} & \textbf{28.73}
                   & \textbf{4.07} & \textbf{20.81} & \textbf{37.10} \\
    \bottomrule
  \end{tabular}
  \caption{Prompt-based baseline performance across three validation splits.  Best scores per column are \textbf{bold}; second–best are \underline{underlined}.}
  \label{tab:prompt_baseline_results}
\end{table*}

Table~\ref{tab:prompt_baseline_results} shows the performance of each prompt for the different validation sets in our experiments. We observe that prompt 3 is the best overall performer and, therefore, was included in our main results. 

\section{Dataset Statistics}
\label{sec:data_stats}
\begin{table}[H]
  \centering
  \tiny
  \begin{tabular}{lrrrr}
    \toprule
    \textbf{Dataset} & \textbf{Mean} & \textbf{Min} & \textbf{Max} & \textbf{Std.\ Dev.} \\
    \midrule
    \texttt{\textsc{VerbosityCTRL}$\_\text{train}$}  & 95.35  & 1  & 200 & 57.08 \\
    \texttt{\textsc{VerbosityCTRL}$\_\text{val}$}    & 92.85  & 1  & 199 & 55.05 \\
    \texttt{\textsc{VerbosityCTRL}$\_\text{range}$}  & 110.00 & 20 & 200 & 57.45 \\
    \texttt{Alpaca\_LI}            & 108.60 & 1  & 200 & 57.98 \\
    \bottomrule
  \end{tabular}
  \caption{Word‐count statistics for the datasets used in our experiments.}
  \label{tab:dataset_statistics}
\end{table}

Table~\ref{tab:dataset_statistics} presents the dataset statistics of the training and validation sets of \dataset as well as the Alpaca\_LI validation set used in our experiments. 

\section{Training}
\label{sec:training}
All models were loaded in using BF16 and FlashAttention 2 (other than gemma-7B-it which was incompatible with FlashAttention 2.) All models utilized a cosine scheduler with warmup with a \texttt{warmup\_ratio} of 0.03 for training along with a \texttt{weight\_decay} of 0.001, \texttt{max\_grad\_norm} of 0.3, \texttt{per\_device\_train\_batch\_size} of 1, and \texttt{gradient\_accumulation\_steps} of 4 where the \texttt{per\_device\_train\_batch\_size} was set to a low value due to memory constraints. The \texttt{num\_train\_epochs} and \texttt{learning\_rate} for each model was determined through a hyperparameter search as detailed in Appendix~\ref{sec:hyperparameter_search}. 

\subsection{Hyperparameter Search}
\label{sec:hyperparameter_search}

We perform a hyperparameter grid search for each tested model for the \texttt{epochs} and \texttt{learning\_rate} hyperparameters. We created a random subset of \dataset$_{\text{train}}$ with 10000 samples that was used for the grid search. We searched over \texttt{epoch} values of $[3,5,7]$ and \texttt{learning\_rate} values of $[5\times10^{-6}, 1\times10^{-5}, 5\times10^{-5}, 1\times10^{-4}]$ and evaluated performance on \dataset$_{\text{val}}$ to determine the \texttt{epoch} and \texttt{learning\_rate} for each model.

\subsection{Hardware}
\label{sec:Hardware}
All experiments were conducted on a single node setup using A6000, L40, and L40S. All reported results are from full model fine-tuning done using DeepSpeed Stage 3 using 8 GPUs. 

\section{Validation}
\label{sec:validation}
All inference was conducted using a \texttt{temperature} of 0 and \texttt{batch\_size} of 4.

\section{Detailed Results Analysis}
\label{sec:detailed_analysis}
\paragraph{Enhanced Performance Across Models.} 
Our \method approach significantly improved word count control across evaluation metrics. On \dataset$_{\text{val}}$, average CPR increased from 5.1\% to 9.4\%, with CPR@0.05 rising from 12.3\% to 38.2\% and CPR@0.1 from 21.4\% to 62.6\%. Similar trends appeared on \dataset$_{\text{range}}$, where average CPR rose from 1.6\% to 4.8\%, with CPR@0.05 and CPR@0.1 improving by 27.1 and 46.3 percentage points, respectively. These benefits extended to out-of-distribution data (Alpaca-LI), demonstrating COMPASS enhances word count control in both in-distribution and out-of-distribution contexts, with the only exception being \metric for LLaMA-3-8B-IT on \dataset$_{\text{val}}$.

\paragraph{Threshold-Specific Performance Gains.}
\method yielded the most substantial improvements at the \metric@0.05 tolerance threshold. On \dataset$_{\text{val}}$, LLaMA-3-8B-IT's \metric@0.05 increased by 23.10 percentage points (22.70\% to 45.80\%), Gemma-7B-IT by 24.00 points (4.90\% to 28.90\%), and Qwen-1.5-7B by 30.40 points (9.40\% to 39.80\%). While improvements occurred across all thresholds, gains were most pronounced at \metric@0.05, suggesting our method effectively refines near-miss output lengths. This pattern remained consistent across validation sets, including out-of-distribution data. The diminishing relative gains at \metric@0.1 indicate that base models already capture many approximate matches at looser tolerances.

\paragraph{Analysis Across Models.}
We observed an inverse relationship between baseline capability and relative improvement with \method. LLaMA-3-8B-IT demonstrated the highest baseline control on \dataset$_{\text{val}}$ (9.80 \metric), followed by Qwen-1.5-7B (4.60 \metric) and Gemma-7B-IT (0.80 \metric). However, initially weaker models achieved the most dramatic gains: Gemma-7B-IT showed over 10$\times$ improvement at \metric (0.80 to 8.90) while LLaMA-3-8B-IT made slight sacrifices at exact-match for substantial gains at relaxed thresholds. These improvements generalized across both \dataset$_{\text{range}}$ and out-of-distribution Alpaca-LI evaluation sets, confirming the method's robustness and effectiveness with initially weaker models.

\section{Analysis Across Different Ranges of Control.}
\label{sec:diff_ranges}
\begin{figure*}[!t]
  \centering
  \includegraphics[width=\textwidth]{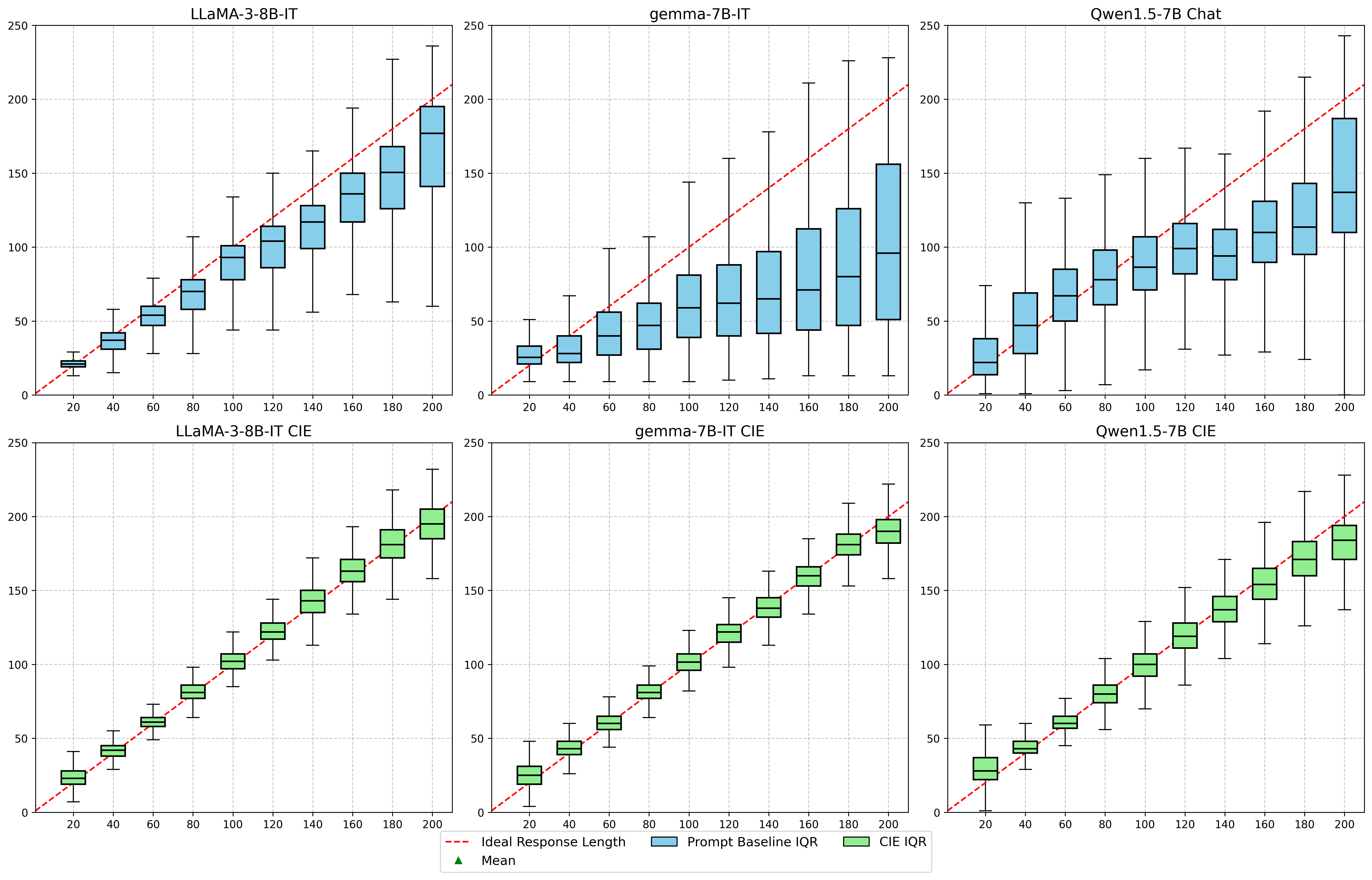}
  \caption{Box-and-whiskers plot of prompt-based length conditioning (top row) and \method (bottom row) on \dataset$_{\text{range}}$. Each box plot contains 1000 datapoints as \dataset$_{\text{range}}$ contains the same 1000 instructions with 10 different conditioned word counts. }
  \label{fig:control_comp}
  \vspace{-15pt}
\end{figure*}

Figure~\ref{fig:control_comp} shows performance on \dataset$_{\text{range}}$, where we created 10 variants for each instruction in \dataset$_{\text{val}}$ with target word counts from 20 to 200. Ideally, each box plot would show minimal interquartile range (IQR) and whiskers, with means aligned along $y = x$. For the prompt baseline, box plot widths and whisker lengths increase with target word counts, indicating greater variance at higher targets. This confirms our previous analysis: LLaMA-3-8B-IT demonstrates the strongest word count adherence, followed by Qwen1.5-7B-IT and Gemma-7B-IT. In contrast, \method variants exhibit dramatically tighter distributions across all target counts, with small IQRs and whisker lengths even at 200 words. Mean responses (green triangles) lie almost exactly on the $y = x$ line, reflecting highly accurate calibration of output lengths. The model performance ranking remains consistent, with \method enforcing word count constraints far more precisely than the prompt baseline.

\section{Win Rates}
\label{sec:win_rates}

\lstset{
  basicstyle=\ttfamily\footnotesize,
  frame=single,
  backgroundcolor=\color{white},
  breaklines=true,
  captionpos=b,
  columns=fullflexible,
  keepspaces=true,
}

To gauge whether \method leads to language generation degradation, we calculate win rates between the generation from \method models and the ``best'' prompt baseline from Appendix~\ref{sec:baseline_prompt}. To reduce order bias when doing LLM-as-judge, we randomize the order of the \method output and the ``best'' prompt baseline output. The model used for judging was GPT-4 and the prompt used is shown in Listing~\ref{lst:win_rate_prompt}. 

\subsection{Win Rates Analysis}
\label{sec:win_rate_analysis}
LLaMA-3-8b-IT demonstrates this most clearly with 77.6\% of evaluations resulting in either wins (22.5\%) or ties (55.1\%), versus 22.4\% losses. Similarly, Gemma-7B-IT shows strong capability preservation with combined 67.9\% for wins and ties (31.3\% and 36.6\% respectively) versus 32.1\% losses. Even Qwen1.5-7B maintains competitive performance with 64.0\% wins or ties (34.9\% and 29.1\% respectively) compared to 36.0\% losses. These findings suggest \method consistently maintains or improves language model instruction following performance relative to the response length conditioned baseline.

\clearpage
\phantomsection
\noindent
{\small
\begin{minipage}[t]{\textwidth}
\phantomsection
    \begin{tcolorbox}[colback=white!95!gray,colframe=gray!50!black,rounded corners,label={box:rubrics}, title={Rubric Outline Example for Expected Action Used to Guide Generation of Examples.}]
Please act as an impartial judge and evaluate the quality of the responses provided by two AI assistants to the user question displayed below. Your evaluation should consider factors such as the helpfulness, relevance, accuracy, depth, creativity, and level of detail of their responses. \\
\\
Begin your evaluation by comparing the two responses and provide a short explanation. Avoid any position biases and ensure that the order in which the responses were presented does not influence your decision. Do not allow the length of the responses to influence your evaluation. Do not favor certain names of the assistants.\\
\\
IMPORTANT: If both assistants provide reasonably adequate answers that address the user's question - even if one might be slightly better in some aspects - you should declare a tie. Only declare a clear winner when one response is substantially superior to the other.\\
\\
Be as objective as possible. Provide a justification for your selection. Your response must end in the format "Therefore the winner is ..." and output your final verdict by strictly following this format: "[[A]]" if assistant A is better, "[[B]]" if assistant B is better, and "[[C]]" for a tie.\\
\\
\textbf{Question:} {prompt} \\
\\
\textbf{Answer A:} {answer\_a} \\
\\
\textbf{Answer B:} {answer\_b}
    \end{tcolorbox}
\label{lst:win_rate_prompt}
\end{minipage}}

\clearpage

\end{document}